# CLASSIFICATION OF LULC CHANGE DETECTION USING REMOTELY SENSED DATA FOR COIMBATORE CITY, TAMILNADU, INDIA

**Y.Babykalpana**

Research Scholar, Anna University, Coimbatore, India.

Dr. K.ThanushKodi

Director, Akshya College Of Engineering And Technology, Coimbatore, India

*ABSTRACT*

Maps are used to describe far-off places . It is an aid for navigation and military strategies. Mapping of the lands are important and the mapping work is based on (i). Natural resource management & development (ii). Information technology ,(iii). Environmental development ,(iv). Facility management and (v). e-governance.

The Landuse / Landcover system espoused by almost all Organisations and scientists, engineers and remote sensing community who are involved in mapping of earth surface features, is a system which is derived from the united States Geological Survey (USGS) LULC classification system. The application of RS and GIS involves influential of homogeneous zones, drift analysis of land use integration of new area changes or change detection etc.,National Remote Sensing Agency(NRSA) Govt. of India has devised a generalized LULC classification system respect to the Indian conditions based on the various categories of Earth surface features , resolution of available satellite data, capabilities of sensors and present and future applications.

The profusion information of the earth surface offered by the high resolution satellite images for remote sensing applications. Using change detection methodologies to extract the target changes in the areas from high resolution images and rapidly updates geodatabase information processing.Traditionally, classification approaches have focused on per-pixel technologies.Pixels within areas assumed to be automatically homogeneous are analyzed independently. These new sources of high spatial resolution image will increase the amount of information attainable on land cover. Significance is that the data can be acquired by our eyes and the energy can be analyzed. But satellites are capable of collecting data beyond the visible band also

However, the traditional method of change detection are not suitable for high resolution remote sensing images. To overcome the limitations of traditional pixel-level change detection of high resolution remote sensing images, based on georeferencing and analysis method, this paper presents a **unsullied** way of multi-scale amalgamation for the high resolution remote sensing images change detection. Experiment shows that this method has a stronger advantage than the traditional pixel-level method of high resolution remote sensing image change detection.

## 1. Introduction

A variety of change detection methods have beendeveloped now a days. Some of the most common methods are (i). image deferencing (2). Principal component analysis, (3). Post-classification comparison, (4). Change vector analysis (5). Thematic change analysis.

Traditionally, classification approaches have focused on per-pixel technologies. [1]Pixels within areas assumed to be automatically homogeneous are analyzed independently. These new sources of high spatial resolution image will increase the amount of information attainable on land cover[2].

The traditional method of storing , analyzing and presenting spatial data is the map.The accuracy of interpretation for the several categories should be about equal. For this paper 2003,2008 and 2009 data have been taken for classifying.

The classification system should be applicable over extensive areas.The map is the features of the earth drawn to scale.Categories should be divisible into more detailed sub categories that can be obtained from large-scale imagery or ground surveys.

Aggregation of categories must be possible. Compression with future land use land over data should be possible.Multiple uses of land should be remitted when possible.

For efficient planning and management, the classified data in a timely manner, in order to get the classified data of the ground; satellites are the best resources to provide the data in a timely manner

**Fig 1.1** . Overview of Remotesensing Process for Landuse and Landcover

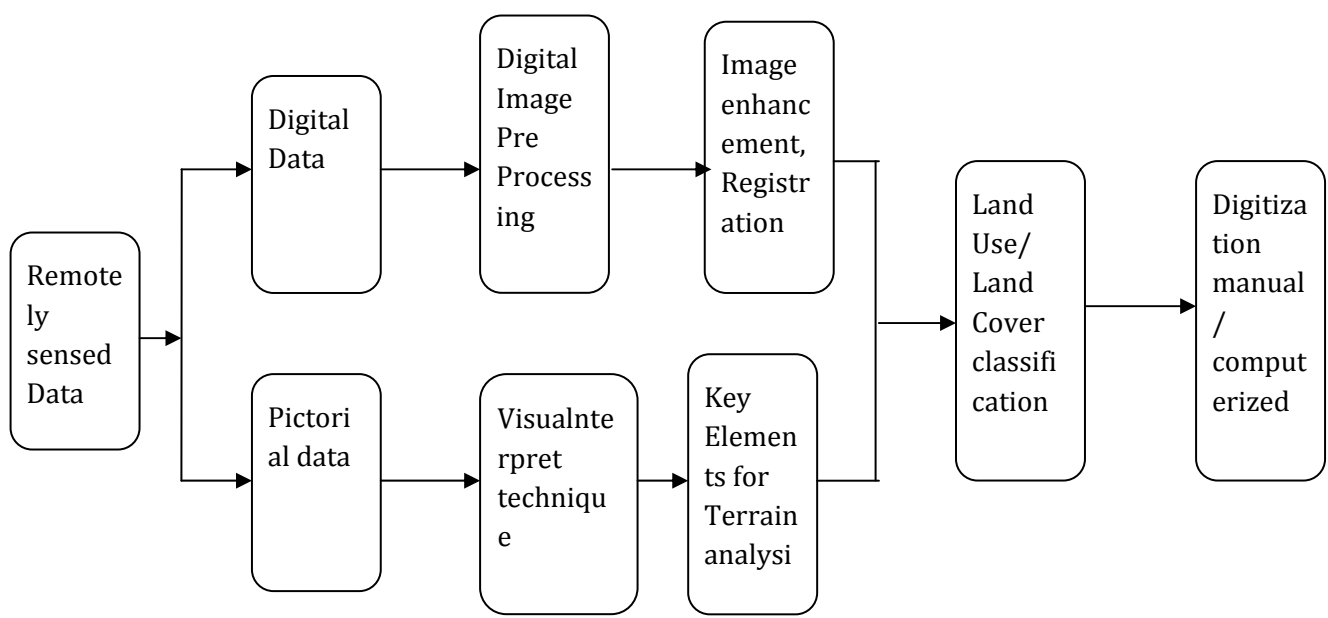

Significance is that the data can be acquired by our eyes and the energy can be analyzed. But satellites are capable of collecting data beyond the visible band also. This will help us to analyze the new things which are not possible in visible band[3].

## 2. Study area

### GENERAL INFORMATION

Area : 105.5 km$^2$ ( 41 sq.m)        Population : 1.5 million (2009 census)

Lattitude N : 10° 10' and 11° 30'     Longitude E : 76° 40' and 77° 30'

Altitude: 43.2 m                      Clothing : Light Cottons

Language Spoken: Tamil

Climate: Tropical Temperature Range (deg C):

Summer: Max 41ºC, Min 38ºC      Winter: Max 32.8ºC Min 20.7ºC   Rainfall: 92.2 mm

### Fig 1.2 Remotely sensed image  - TN

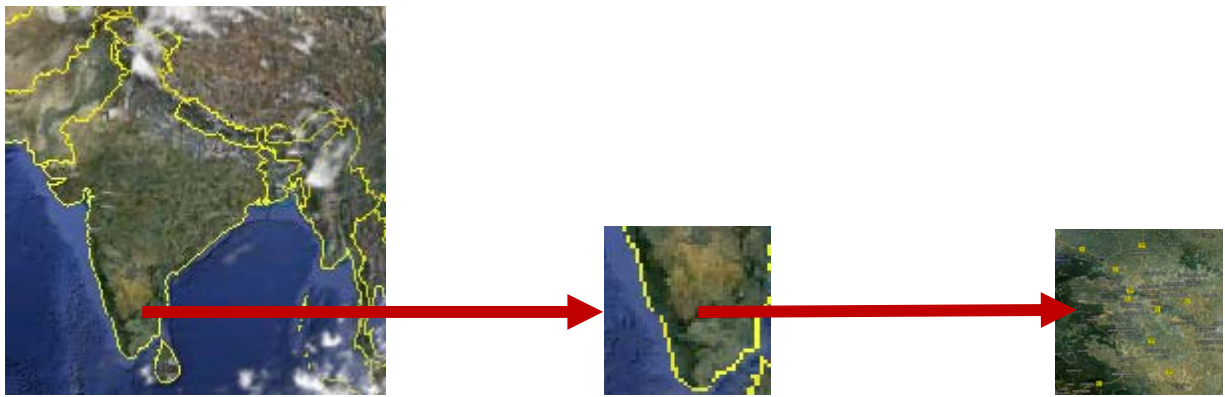

## 3. Methodology

Normally, the process includes edge detection techniques, image acquisition, image enhancement, segmentation classification, data modeling etc.

### 1. Image acquisition

Numerous electromagnetic and some ultra sonic sensing devices frequently are arranged in an array format. CCD sensors are used in digital cameras and other light sensing instruments [8].

## 1.1. Preprocessing

To correct distorted or degraded image data to create a more faithful representation of the original scene, image rectification and restoration process is necked which is always termed as preprocessing.

## Edge detection

An edge is a set of connected pixels that his on the boundary between two regions, Edge detection is performed on the image by the construction of edge detection operators like sobel edge detection, laplacian edge operator etc. For a continuous image f(x,y), where x and y are the row and column coordinates respectively, consider 2D derivatives $\delta y f(x,y)$ and $\delta x f(x,y)$. Two functions can be expressed :

1. Gradient magnitude

$$|\Delta f(x,y)| = \sqrt{(\partial_x f(x,y))^2 + (\partial_y f(x,y))^2}$$

2. Gradient orientation

$$\angle \Delta f(x,y) = ArcTan(\partial_y f(x,y) / \partial_x f(x,y))$$

Local maxima of the gradient magnitude identify edges in f(x,y). The first derivative achieves a maximum and the second derivative is zero. For this reason, an alternative edge detection strategy is to locate zeros of the second derivatives of f(x,y) [2].

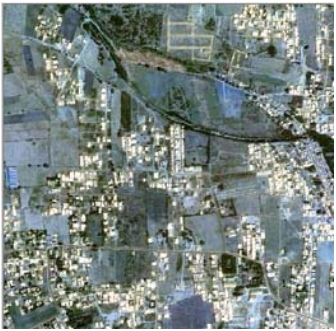

Fig 1.3. edge detection sample o/p

## 2. Image segmentation technique

Image segmentation is the partition and pick-up of the homogeneous regions of image. In the results of segmentation, the consistancy of gray the smoothing of boundary and the connectivity are fulfilled. The classical method of segmentation is the spatial cleaning based on the measurement space[1].

Image segmentation is crucial processing procedure for the classifications and feature extraction of high resolution remote sensing image[5]. The segmentation result is able to sway the effect of subsequent processing. At present the main image segmentation methods are

1. Thresh-hole based

2. Edge-band

3. Region-based

The edge-based segmentation is taken into account which is namely grounded on discontinuity of gray-level in imagery. The image is segmented by the edge of the different homogenous areas[4].

Adopting this method, the accuracy of edge positioning is high whereas the consecutive edge composed of a serial of unique pixels cannot produced, so a sequent process including bulky the detected edge points should he requisite.

## 4. Clustering technique

There are numerous clustering algorithms that can be used to determine the natural spectral groupings present in a data set. One common form of clustering, called the "k-means" approach accepts from the analyst the no. of clusters to be located in the data[6].

The algorithms then arbitrarily "seeds" or locates that number of clusters centers in the multidimensional measurement space[8]. Each pixel in the image is then assigned to the cluster whose arbitrary mean vector is closet.

4.1. G-Static for measuring high/low clustering

It cannot tell whether the clustering is made of high values or low values. This led to the use of G-Statistic by Moran[6], which can separate clusters of high values from clusters of low values. The general G-statistic based on a specified distance, d, is defined by,

$G(d) = \sum\sum w_{ij}(d) x_i x_j / \sum\sum x_i x_j, \ i \neq j$

Where $x_i$ is the value at location i, $x_j$ is the value at location j if j is within d of i, and $w_{ij}(d)$ is the spatial weight. The weight can be based on some weighted distance.

The expected value of G(d) is

$E(G) = \sum\sum w_{ij}(d) / n(n-1)$

E(G) is typically very small value when n is large.

## 5. Unsupervised classification

The family of classifiers involves algorithms that examine etc unknown pixels in an image and aggregate them into a number of classes based on the natural groupings or clusters present in the image values.

The basic premise is that values with in a given cover type should be close together in the measurement space; where as data in different classes should be comparatively well separated.

The classification algorithm is designed to automatically said dense regions within the n-dimensional hyper spectral data cloud[7]. The algorithm is based on the well-known observation that spectra of large, distinct land covers tend to cluster around a mean spectrum.

This is the basis for unsupervised classification parcel on cluster analysis[2]. The pixel density around the mean spectrum depends on the spectral variability of the land cover and the areas extent of the land cover.

5.1. Geo spatial Data

A GIS represents spatial features on the Earth's surface as map features on a plane surface. This transformation involves 2 main issues[6]: (i). spatial reference system (ii). Data model.

The spatial features are based on a geographic coordinat system with longitude ( $77^\circ$ N) and Lattitude ( $11^\circ$ E) values where as the locations of map features are based on plane coordinate system with x,y-coordinates.

The data model defines how spatial features are represented in a GIS. (i). Vector model uses points and their x,y-co-ordinates to construct spatial features of points , lines, and areas.

(2). The raster data model uses[6] a grid and grid cells to represent the spatial variation of a feature. The two data models differ in concept: vector data are ideal for representing dicrete features; raster data are better suited for continous features where in this paper we used the raster.

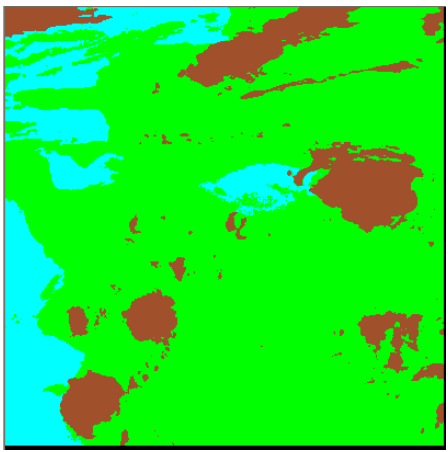 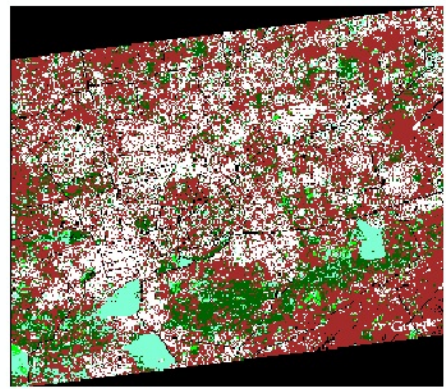

**Fig. 5.1. Geo Referencing images**

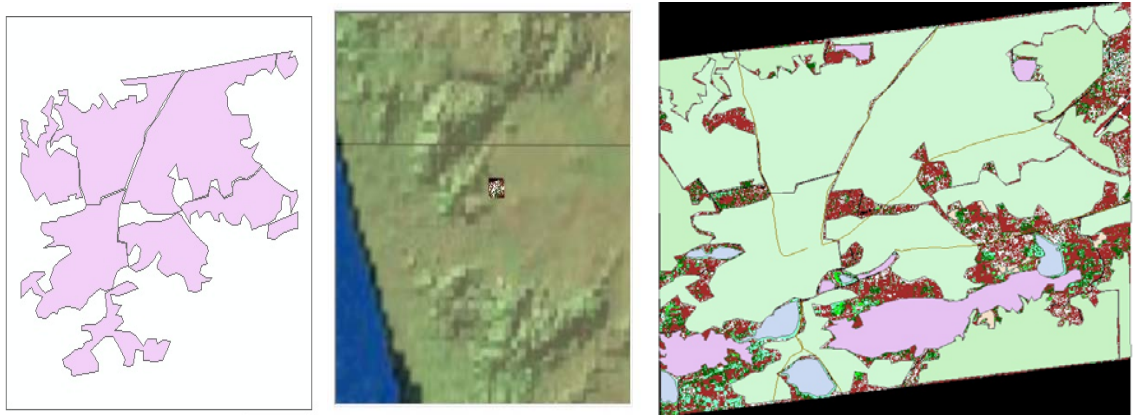

**Fig. 5.2. Classification of Coimbatore city**

## 6. Conclusion

**Sample data**

| Category | Hectares | Land Use |
|---|---|---|
| 1 | 135 | Water |
| 2 | 210 | Aquatic Vegetation |
| 3 | 90 | Urban/Edge |
| 4 | 190 | Grassland |
| 5 | 169 | Bare Soil |
| 6 | 270 | Forest |
| - | 1064 | Total |

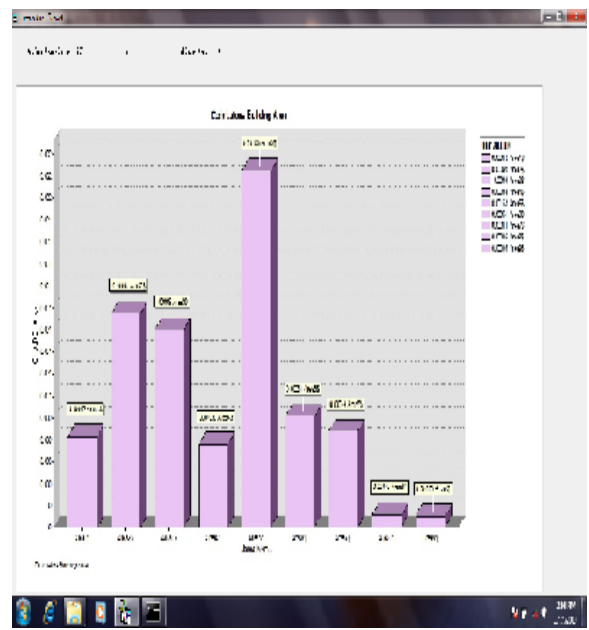

**Fig 6.1. Report generated on Land Use**     **Fig. 6.2 Urban Area Report**

      The remote sensing data have been analyzed to fixed the land cover classification of our city, and to know how the use of land changes according to time and also performed the temporal analysis to analyze[3] all these things, the unsupervised classification method is used.This is very fast and useful analysis method. It is widely used for the crops[10] classification in the world and this classification method is used for land cover and land use because vegetation components are important in the images[9].The basic axis is also to preserve the greenery of the city for the healthy environment.

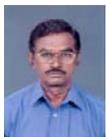 Dr.K.ThanushKodi B.E., M.Sc(Engg).,Ph.D., is currently working as the director, Akshya college of Engineering.He has more than 40 years of experience as a teaching facuty. And more than 5 years as Principal of various engineering colleges. He is guiding more than 30 scholars. He is the respected syndicate member of Anna university,Coimbatore,India. He is serving as Chairman of Board of studies, Dept. of EEE at Anna University, Coimbatore. He is also a member of Board of Studies, Academic Councils of various universities across Tamilnadu. He has vast teaching, research and administrative experience in various Government and Private Engineering Colleges for more than 3 decades.

Y. Baby Kalpana received the B.E. and M.E. degrees in Computer Science and Engineering from Tamilnadu college of engineering in 1997 and 2006, respectively. She is working as a Lecturer in the Department of computer Science and Engineering , Tamilnadu college of Engineering,Coimbatore,India. She is one of the Board of Study members, SUITS,IECD, Bharathidasan university, Trichy,India. She is now pursuing her Ph.D degree programme under the excellent guidance of Dr. K. Thanushkodi ,Director , Akshya College of engineering and Technology, Coimbatore,India.